\documentclass[a4paper,fleqn]{cas-dc}

\usepackage[numbers]{natbib}
\usepackage{xcolor}
\usepackage{verbatim}
\usepackage{tcolorbox}
\usepackage{hyperref}
\hypersetup{
colorlinks=true,
linkcolor=blue,
anchorcolor=blue,
urlcolor=blue,
citecolor=blue,
}
\usepackage{amsmath}

\def\tsc#1{\csdef{#1}{\textsc{\lowercase{#1}}\xspace}}
\tsc{WGM}
\tsc{QE}
\tsc{EP}
\tsc{PMS}
\tsc{BEC}
\tsc{DE}
\begin{document}
\let\WriteBookmarks\relax
\def\floatpagepagefraction{1}
\def\textpagefraction{.001}
\shorttitle{Improving Stack Overflow question title generation with copying enhanced CodeBERT model and bi-modal information}
\shortauthors{Fengji Zhang et al.}

\title [mode = title]{Improving Stack Overflow question title generation with copying enhanced CodeBERT model and bi-modal information}                      
%\tnotemark[1,2]

%\tnotetext[1]{This document is the results of the research
%   project funded by the National Science Foundation.}

\author{Fengji Zhang\textsuperscript{\textit{a,b}}}[style=chinese]
\ead{zhangfengji@whu.edu.cn}
\address{\textsuperscript{\textit{a}}School of Computer Science and Artificial Intelligence, Wuhan University of Technology, Wuhan, China}
\address{\textsuperscript{\textit{b}}School of Computer Science, Wuhan University, Wuhan, China}

\author{Xiao Yu\textsuperscript{\textit{a,c,d}}}[style=chinese]
\ead{xiaoyu@whut.edu.cn}
\address{\textsuperscript{\textit{c}}Wuhan University of Technology Chongqing Research Institute, Chongqing, China}
\address{\textsuperscript{\textit{d}}Sanya Science and Education Innovation Park of Wuhan University of Technology, Sanya, China}

\corref{mycorrespondingauthor}
\cormark[1]

\author{Jacky Keung\textsuperscript{\textit{e}}}[style=chinese]
\ead{jacky.keung@cityu.edu.hk}
\address{\textsuperscript{\textit{e}}Department of Computer Science, City University of Hong Kong, Hong Kong, China}

\author{Fuyang Li\textsuperscript{\textit{a,c}}}[style=chinese]
\ead{fyli@whut.edu.cn}

\corref{mycorrespondingauthor}
\cormark[1]

\author{Zhiwen Xie\textsuperscript{\textit{b}}}[style=chinese]
\ead{xiezhiwen@whu.edu.cn}

\author{Zhen Yang\textsuperscript{\textit{e}}}[style=chinese]
\ead{zhyang8-c@my.cityu.edu.hk}

\author{Caoyuan Ma\textsuperscript{\textit{b}}}[style=chinese]
\ead{macaoyuan@whu.edu.cn}

\author{Zhimin Zhang\textsuperscript{\textit{b}}}[style=chinese]
\ead{zhangzhimin@whu.edu.cn}

\cortext[cor1]{Corresponding author}

\begin{abstract}
%\boldmath
\renewcommand{\textbf{Context:}} Stack Overflow is very helpful for software developers who are seeking answers to programming problems. Previous studies have shown that a growing number of questions are of low quality and thus obtain less attention from potential answerers. Gao et al. proposed an LSTM-based model (i.e., BiLSTM-CC) to automatically generate question titles from the code snippets to improve the question quality. However, only using the code snippets in the question body cannot provide sufficient information for title generation, and LSTMs cannot capture the long-range dependencies between tokens.\\
\textbf{Objective:} This paper proposes CCBERT, a deep learning based novel model to enhance the performance of question title generation by making full use of the bi-modal information of the entire question body.\\
\textbf{Method:} CCBERT follows the encoder-decoder paradigm and uses CodeBERT to encode the question body into hidden representations, a stacked Transformer decoder to generate predicted tokens, and an additional copy attention layer to refine the output distribution.
Both the encoder and decoder perform the multi-head self-attention operation to better capture the long-range dependencies. This paper builds a dataset containing around 200,000 high-quality questions filtered from the data officially published by Stack Overflow to verify the effectiveness of the CCBERT model. \\
\textbf{Results:} CCBERT outperforms all the baseline models on the dataset. Experiments on both code-only and low-resource datasets show the superiority of CCBERT with less performance degradation. The human evaluation also shows the excellent performance of CCBERT concerning both readability and correl
ation criteria. \\
\textbf{Conclusion:} CCBERT is capable of automatically capturing the bi-modal semantic information from the entire question body and parsing the long-range dependencies to achieve better performance. Therefore, CCBERT is an effective approach for generating Stack Overflow question titles.

\end{abstract}

\begin{keywords}
Stack Overflow \sep Title generation \sep
\sep Copy mechanism \sep CodeBERT
\end{keywords}

\maketitle
\section{Introduction}
Stack Overflow (SO) is one of the most thriving communities where software developers can seek answers to programming problems from peers. The open-data policy of SO has been attracting intense research interests \cite{Chakraborty2021HowDD, Rubei2020PostFinderMS, Uddin2020MiningAU, mondal2021early, MONTANDON2021106429, TAHIR2020106333}. The recent study of Mondal et al. \cite{mondal2021early} shows that a growing number of open questions in SO remain unanswered, partly because some developers fail to write high-quality questions. The SO community has given many practical writing suggestions in the official tutorial\footnote{\url{https://stackoverflow.com/help/how-to-ask}} to tackle this problem, and researchers have also made great efforts to help improve question quality in many ways \cite{wang2018users, wang2019sotagrec, gao2020generating}.

Previous studies \cite{arora2015good, calefato2018ask, correa2013fit, yao2013want} in SO have demonstrated the importance of question titles to the overall quality of questions. Recently, Gao et al. \cite{gao2020generating} for the first time proposed an approach of automatically generating question titles from given code snippets. They used the \underline{BiLSTM}-\underline{CC} model, which is a \underline{Bi}-directional \underline{L}ong \underline{S}hort-\underline{T}erm \underline{M}emory network incorporated with the \underline{C}opy \cite{Gu2016IncorporatingCM} and \underline{C}overage \cite{Tu2016ModelingCF} mechanism to generate titles from code snippets mined in corresponding question bodies. Despite the encouraging performance, we argue that LSTMs may lack the ability to parse long-range dependencies according to Khandelwal et al. \cite{khandelwal2018sharp}.
In addition, developers are not recommended to write only source code as questions in the SO community. Since a question body usually consists of the bi-modal content (i.e., text descriptions and code snippets), the information that developers infer from code snippets without surrounding contexts can be broken and misleading.

In this paper, we redefine the task proposed by Gao et al. \cite{gao2020generating} to \textbf{T}itle \textbf{G}eneration from the \textbf{E}ntire \textbf{Q}uestion \textbf{B}ody, namely \textbf{TGEQB}. We formulate this task as an abstractive summarization problem, and also propose our \textbf{CCBERT} model, which combines the \textbf{C}opy mechanism \cite{Gu2016IncorporatingCM} to handle rare tokens and the pre-trained \textbf{C}ode\textbf{BERT} \cite{Feng2020CodeBERTAP} model to parse bi-modal content. We follow the encoder-decoder paradigm and use CodeBERT to encode question bodies into hidden representations, a stacked Transformer decoder to generate predicted tokens, and an additional copy attention layer to refine the output distribution. Our encoder and decoder perform the multi-head self-attention operation, which helps CCBERT better capture the long-range dependencies than LSTMs.

To verify the effectiveness of our model, we conduct the empirical study by raising the following Research Questions (RQs):

\textbf{RQ-1 Does our CCBERT model outperform the baseline models?}
We build a large-scale dataset $\mathrm{Data_{exp}}$ with around 200,000 high-quality questions filtered from the data\footnote{\url{https://archive.org/download/stackexchange}} officially published by Stack Overflow in December 2020, which contains all the historical questions from 2008 to 2020. We employ BLEU and ROUGE as the evaluation metrics and choose four baseline models (i.e., TF-IDF \cite{Luhn1958TheAC}, BiLSTM \cite{Schuster1997BidirectionalRN}, BiLSTM-CC \cite{gao2020generating}, and BART \cite{Lewis2020BARTDS}). Experimental results show that CCBERT outperforms all the baseline models regarding all the metrics.

\textbf{RQ-2 What is the advantage of using the bi-modal information of the entire question body?}
We build a code-only dataset and choose BiLSTM-CC to compare with our model. Experimental results show that applying bi-modal information greatly boosts both models' performance, where CCBERT still outperforms BiLSTM-CC.

\textbf{RQ-3 How effective is our CCBERT model under low-resource circumstances?}
We build three new train sets sized of 98,909 ($\mathrm{Data_{exp}/2}$), 49,454 ($\mathrm{Data_{exp}/4}$), and 24,727 ($\mathrm{Data_{exp}/8}$) to train the CCBERT and BiLSTM-CC models. According to the experimental results, our CCBERT model shows significant superiority under low-resource circumstances compared with BiLSTM-CC. 

\textbf{RQ-4 How much influence does interrogative constraint have on model training?}
We follow Gao et al. \cite{gao2020generating} to apply the interrogative constraint when building $\mathrm{Data_{exp}}$, which may make it easier to train our models. However, our model can be biased because of this. We build another dataset $Data_{exp+}$ to quantify the influence, and results show that dropping the interrogative constraint will lead to a decline in the performance of both CCBERT and BiLSTM-CC models.

\textbf{RQ-5 How effective is our CCBERT model under human evaluation?}
Automated evaluation is not always trustworthy. So, we perform a more human-centered evaluation to investigate the overall quality of the titles generated by our models. Results show that our CCBERT performs much better than TF-IDF and BiLSTM-CC concerning the correlation criteria but a little bit worse than human written titles retrieved by TF-IDF concerning the readability criteria under human evaluation.

The contributions of this paper are as follows:
\begin{itemize} 
	\item We introduce a new task named TGEQB, which is to generate high-quality titles from the entire question body containing bi-modal content to help improve the question quality.
	
	\item We propose a novel model named CCBERT, which combines the copy mechanism and CodeBERT to handle rare tokens and long-range dependencies in the bi-modal context.
	
	\item We have released our dataset and all relevant source code\footnote{\url{https://github.com/zfj1998/SO_Title_Generation}} to facilitate future research and application.
\end{itemize}

We organize the rest of this paper as follows:
Section \ref{motivation} reveals the motivation of our work. Section \ref{proposed approach} introduces the details of our proposed approach. Section \ref{experimental setup} describes the basic setup of our experiment, including the baseline models, evaluation metrics, and model settings. Section \ref{results and analysis} presents the experimental results. Section \ref{related work} introduces the related works. Section \ref{threats to validity} discusses threats to the validity of our work. Finally, we conclude this paper and introduce the future work in Section \ref{conclusion and future work}.

\section{Motivation}
\label{motivation}

\begin{figure}
	\centering
	\includegraphics[scale=.6]{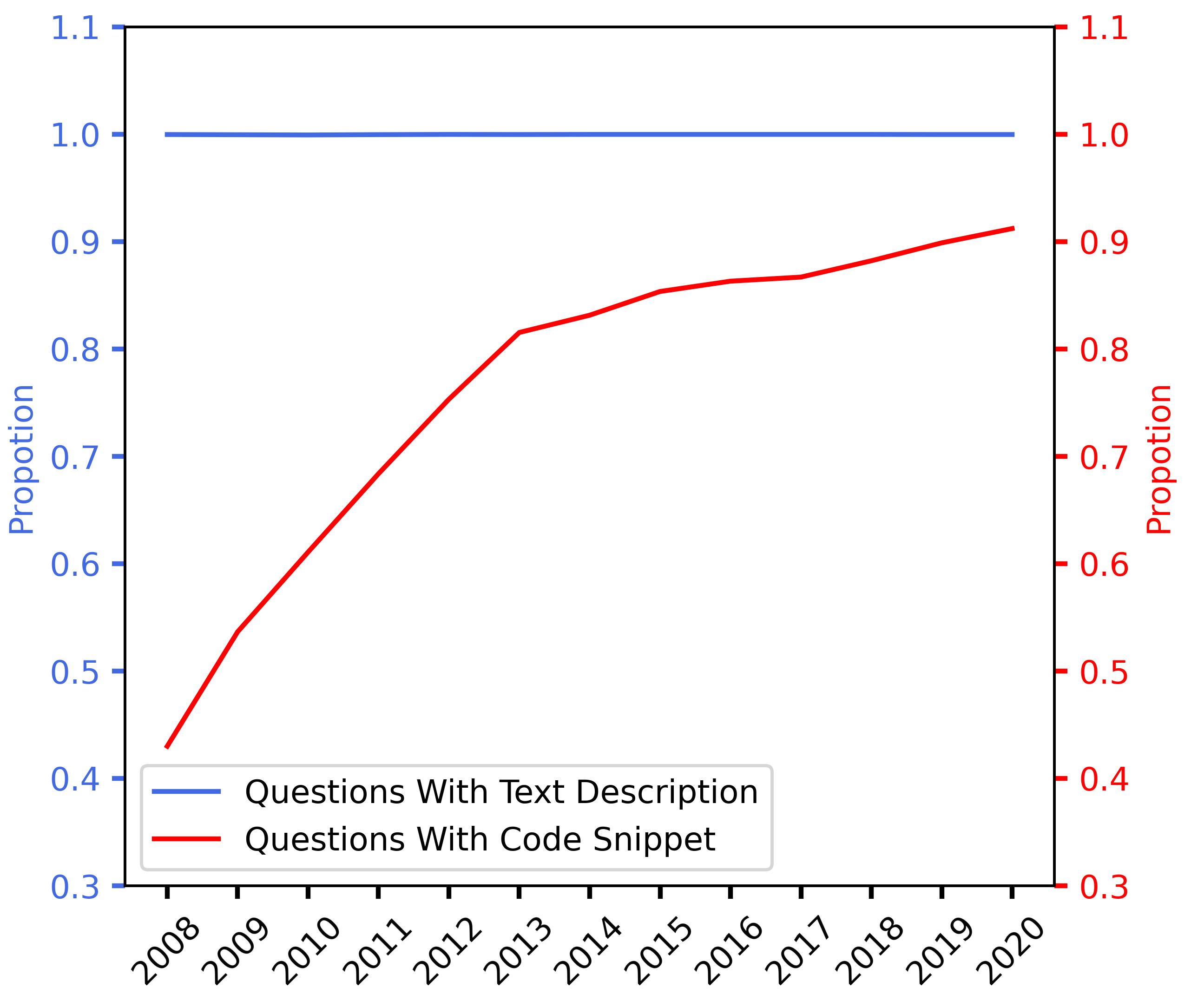}
	\caption{The proportion of questions with text descriptions or code snippets}
	\label{FIG:Fig. 1}
\end{figure}

\begin{figure}
	\centering
	\includegraphics[scale=.75]{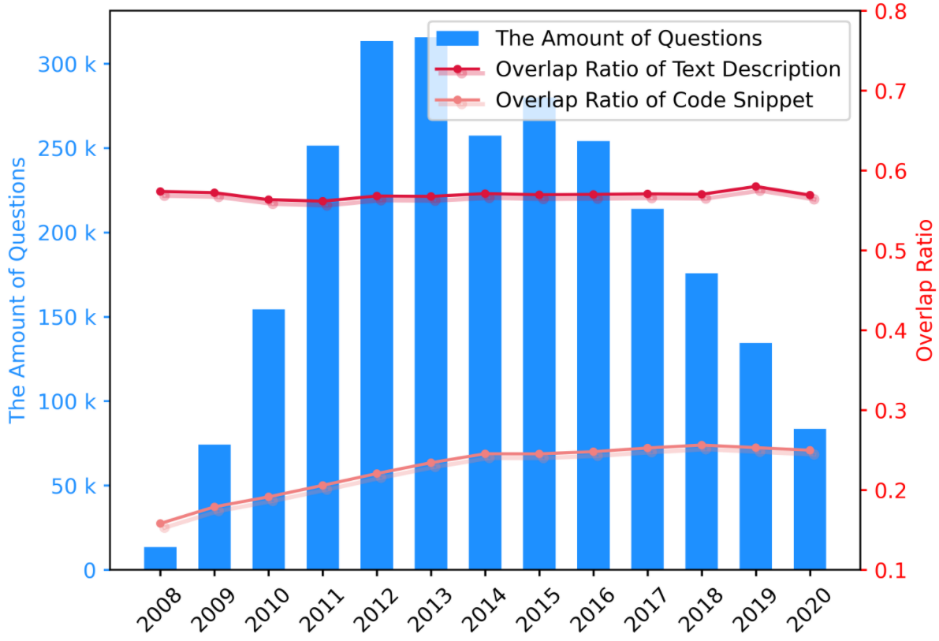}
	\caption{The overlap between titles and code snippets/text descriptions}
	\label{FIG:Fig. 2}
\end{figure}

We share similar user scenarios with Gao et al. \cite{gao2020generating}, where less experienced developers or non-native English speakers may not be able to adequately describe their questions according to the writing rules suggested by SO. We can use an automated data-driven approach to help developers draft high-quality question titles in such circumstances. However, we argue that one should consider both text descriptions and code snippets when writing question titles. We also have concerns about the long-range dependency issues of using the entire question body as input. Therefore, this section aims to investigate the necessity of title generation using bi-modal content and the challenge of long sequence parsing.

\subsection{Importance of bi-modal Content}
Our first step is to get a collection of high-quality samples for statistics. We believe that high-quality question posts should be clear and complete so that developers in the SO community are more likely and willing to answer. But it is a non-trivial task to evaluate the clearness and completeness of a post automatically.
Therefore, we first filter the question posts based on the feedback they received from the SO community:
\begin{enumerate}\label{filter condition}
	\item The question is not closed; 
	\item The question has an accepted answer; 
	\item The question gets more than one vote.
\end{enumerate}
After filtering out the ones that do not meet the above feedback-related constraints, we obtain a collection of 3.2 million question posts that we regard as candidates of high-quality ones.

Intuitively, a programming question should always contain both text and code. Besides, when drafting a new question post in SO, the website will give the three suggestions: \footnote{\url{https://stackoverflow.com/questions/ask}}
\begin{enumerate} \label{writing suggestions}
	\item Summarize the problem;
	\item Describe what you have tried;
	\item Show some code.
\end{enumerate}

We separately count the high-quality candidate questions containing text descriptions and code snippets by year. To be specific, question posts in the source file\footnote{the \emph{stackoverflow.com-Posts.7z} file} are organized in a unified HTML format, so we extract the content wrapped by "<code></code>" tags as code snippets and the rest as text descriptions. We draw a line chart in Fig. \ref{FIG:Fig. 1} to show the statistical results, where the x-axis denotes the year and the y-axis denotes the proportion of questions with text descriptions or code snippets. 
We find that the proportion of high-quality candidates containing text descriptions is almost unchanging (100\%) every year. While the proportion of high-quality candidates containing code snippets has been increasing in recent years, reaching 90\% in 2020.

In addition, we have manually studied a number of high-quality candidates posted recently without code snippets. We find that many of them are not programming-related questions, such as software/platform instructions,\footnote{\url{https://stackoverflow.com/questions/59553413/firebase-storage-image-not-showing}} knowledge Q\&A,\footnote{\url{https://stackoverflow.com/questions/59554837/uml-how-to-model-either-or-both-union-concept}} etc. Some others may put the code in images\footnote{\url{https://stackoverflow.com/questions/59552547/ios-swiftui-how-to-bring-up-extra-actions-like-embed-in-vstack-when-interactin}} or external links.\footnote{\url{https://stackoverflow.com/questions/59552571/how-to-check-if-fixed-width-integers-are-defined}} The above-mentioned question posts are beyond the scope of this preliminary study.

Therefore, we believe that containing the content of both modalities helps improve the clearness and completeness of programming questions in SO. In our study, only the question posts containing bi-modal content and meeting the three feedback-related constraints are considered high-quality.

\subsection{Impact of bi-modal Content on Titles}
\label{impact_bi_modal_on_title}

\begin{table*}
	\centering
	\caption{The criteria used for manually evaluating the titles. The evaluation score of each criteria is between 1 and 4.}
    \label{tbrq_human}
	\begin{tabular}{@{\ \ }l|l|l@{\ }}
		\toprule
		Criteria  & Description & Title Scoring Standard \\ \midrule[0.6pt]
        Readability & 
        \begin{tabular}[c]{@{}l@{}} Ignoring the content, considering the\\ grammaticality and fluency of the title \end{tabular} & 
        \begin{tabular}[c]{@{}l@{}}1. Has too many errors to read and understand \\2. Has minor errors but is readable and understandable \\3. Is very easy to read and understand \\4. Is very expressive and appealing\end{tabular} \\ \midrule[0.4pt]
        Correlation & 
        \begin{tabular}[c]{@{}l@{}} Considering the consistency between\\ the question and the title \end{tabular} & 
        \begin{tabular}[c]{@{}l@{}}1. Is totally missing the point of the question \\2. Is relevant to the main point of the question \\3. Is a good match of the question's points \\4. Is a perfect summary of the question \end{tabular} \\ 
		\bottomrule
	\end{tabular}
\end{table*}

\begin{table*}
	\centering
	\caption{Human evaluation results of the title quality with different overlap ratios with the bi-modal content.The ratios of four different scores and the average score are listed grouped by different criteria.}
    \label{overlap_eval}
	\begin{tabular}{@{\ \ }ccccccc@{\ \ }}
		\toprule
		Criteria   & Overlap Ratio  & Score 1 & Score 2 & Score 3 & Score 4 & Avg Score\\ \midrule[0.6pt]
		\multirow{4}{*}{Readability}
		& \scriptsize 0-0.2    & - & 21.6\% & 77.4\% & 1.0\% & 2.794 \\
		& \scriptsize 0.2-0.4    & - & 7.8\% & 88\% & 4.2\% & 2.964 \\
		& \scriptsize 0.4-0.6    & - & 11\% & 85.6\% & 3.4\% & 2.924 \\
		& \scriptsize 0.6-1.0 & - & 26\% & 73.8\% & 0.2\% & 2.742 \\ \midrule[0.2pt]
		\multirow{4}{*}{Correlation}
		& \scriptsize 0-0.2    & - & 11.2\% & 87\%& 1.8\% & 2.906\\
		& \scriptsize 0.2-0.4    & - & 2.8\% & 86.4\% & 10.8\% & 3.080 \\
		& \scriptsize 0.4-0.6    & - & 2.8\% & 89\%& 8.2\% & 3.054 \\
		& \scriptsize 0.6-1.0    & - & 6.2\% & 90.6\% & 3.2\% & 2.970 \\ \bottomrule
	\end{tabular}
\end{table*}

The title of a question post is always dependent on the overall semantics of the question body. However, it is a non-trivial task to tell the precise impact of the bi-modal content in the question body on writing the title. In this subsection, we conduct a lexical-level statistical experiment and a human evaluation to estimate such impact.

First, we extract the high-quality question posts from 2008 to 2020, which contain bi-modal content and meet the three feedback-related constraints. Then we count the tokens that appear both in the title and text descriptions/code snippets, and draw a line chart in Fig. \ref{FIG:Fig. 2} to demonstrate the average overlap ratios by year. Moreover, we combine Fig. \ref{FIG:Fig. 2} with a bar chart to demonstrate the number of statistical samples per year. Specifically, the x-axis denotes the year, the left y-axis denotes the amount of questions, and the right y-axis denotes the overlap ratios between titles and text descriptions/code snippets.

From Fig. \ref{FIG:Fig. 2}, we may find that the overlap ratios of high-quality questions have been stable since 2014. This may indicate that a certain extent of token overlap between the title and the bi-modal content ensures the title quality. To investigate this issue, we further perform a manual analysis.

We consider two criteria when manually evaluating post titles, either of which can be scored between 1 and 4. We illustrate the detailed descriptions and scoring standards in Table \ref{tbrq_human}. We split the filtered high-quality posts into four categories according to their average overlap ratios between titles and the bi-modal content. And then, we randomly sample 500 question posts from each category. Five independent graduate students who are experienced programmers and familiar with Stack Overflow are invited to rate the titles based on the scoring standards, and each participant is assigned 100 posts per category.
From the evaluation results in Table \ref{overlap_eval}, we can find both the readability and correlation scores degrade when the overlap ratio is too low or too high. According to our participants, when the overlap is too low, the title tends to be vague. When the overlap is too high, the title tends to be the bare error report of a program. Both situations require more effort for the reader to fully understand the question. This suggests that properly borrowing tokens from the bi-modal content makes the title more expressive and matching the points of the question. Therefore, we have reasons to believe that bi-modal content is essential to writing good titles.

\subsection{The Long-range Dependency Issue}

\begin{figure}
	\centering
	\includegraphics[scale=.65]{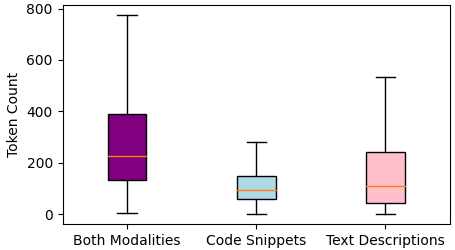}
	\caption{The length distribution of code snippets and text descriptions in the question body}
	\label{FIG:Fig. 3}
\end{figure}

We draw a box plot in Fig. \ref{FIG:Fig. 3} to represent the length distributions of the entire question body, the code snippets, and the text descriptions of the high-quality question posts extracted in Section \ref{impact_bi_modal_on_title}. We can find that the code snippets only occupy less than half of the body content, and the entire content of a question body can be very long, bringing new challenges to our title generation models. Specifically, over 56\% questions have more than 200 tokens in their bodies, and some questions even have 500 tokens and more. The LSTM structure is only capable of using 200 tokens of context on average according to Khandelwal et al. \cite{khandelwal2018sharp}, so we apply Transformer-based pre-trained models to tackle this problem. Later in Section \ref{result}, we will make further comparisons of these two models.

\begin{figure*}
	\centering
	\includegraphics[scale=.26]{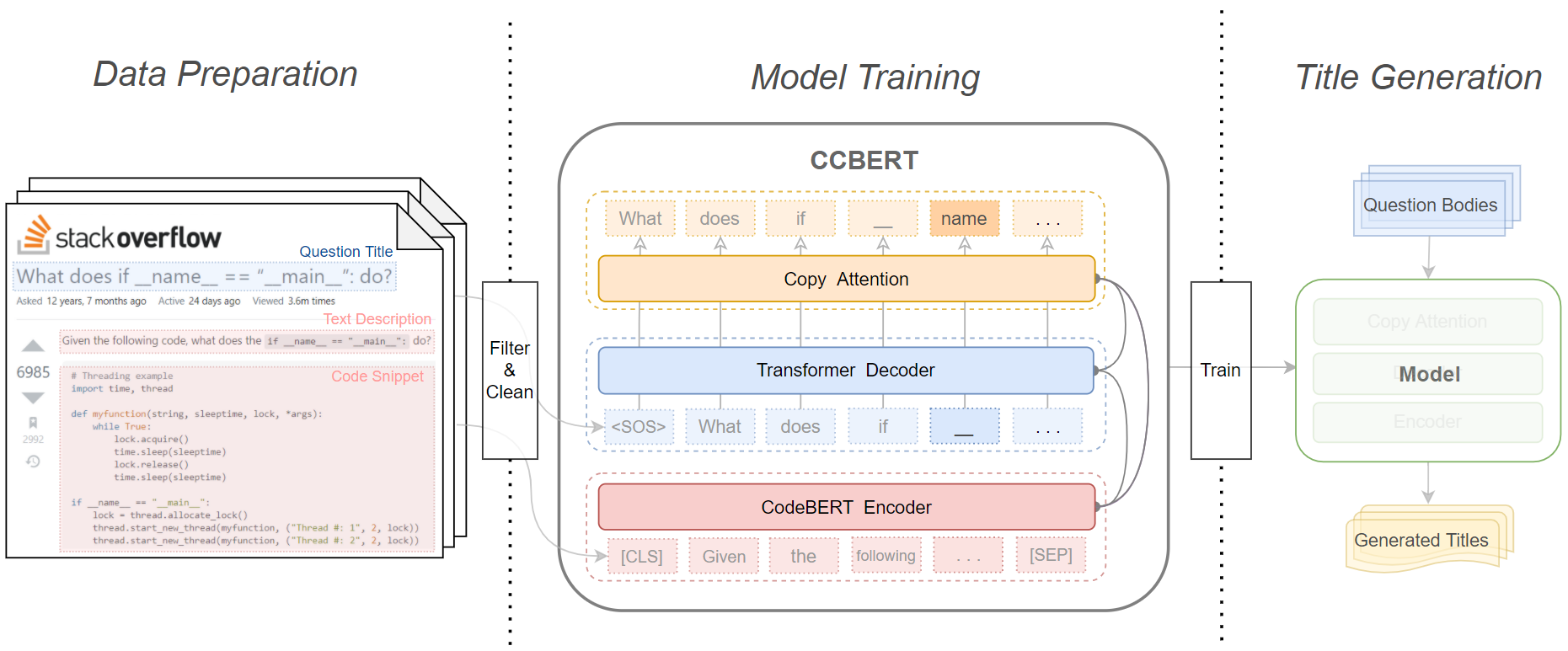}
	\caption{The framework of our approach for Stack Overflow question title generation}
	\label{FIG:Fig. 4}
\end{figure*}

\section{Proposed Approach}
\label{proposed approach}
We aim to help developers write high-quality questions with a better chance to get answers in Stack Overflow. Considering that using only code snippets is not enough to generate high-quality titles, we introduce a new title generation task named TGEQB, which utilizes the bi-modal information of both text descriptions and code snippets in the question body. 
Following the general practice in machine learning studies, the framework of our approach demonstrated in Fig. \ref{FIG:Fig. 4} contains three main steps: data preparation, model training, and validation. We describe the details of our approach in this section, including the data preparation procedures and the detailed architecture of our CCBERT model.

\subsection{Data Preparation}
Further filtering, tokenization, and partitioning are performed on the high-quality question posts extracted in Section \ref{impact_bi_modal_on_title} before we finally get the experimental dataset $Data_{exp}$.

Firstly, considering the vocabulary of our CodeBERT encoder was built on a dataset\footnote{CodeSearchNet \url{https://github.com/github/CodeSearchNet}} concerning only six programming languages: \emph{Java}, \emph{Python}, \emph{JS(JavaScript)}, \emph{PHP}, \emph{Ruby}, and \emph{Go}, we also focus on the questions tagged with these programming languages in this preliminary study. To avoid the influence of noise data, we further filter out posts tagged with other popular languages in SO, including \emph{C\#}, \emph{HTML}, and \emph{C++}. In the end, we find the amounts of filtered \emph{Ruby} and \emph{Go} questions\footnote{There are only approximately 7,000 and 3,500 filtered questions for \emph{Ruby} and \emph{Go}.} far from enough for training and testing, which leaves us the questions tagged with only four programming languages: \emph{Java}, \emph{Python}, \emph{JS}, and \emph{PHP}.

Secondly, we notice that in Gao et al.'s work \cite{gao2020generating}, they only kept the questions containing interrogative keywords:  \emph{how}, \emph{what}, \emph{why}, \emph{which}, and \emph{when} in their titles. While only 1/3 of our filtered high-quality question posts satisfy this constraint. After manually examining our data samples, we find that question titles without interrogatives tend to be more casual and are always incomplete sentences, which undoubtedly brings many difficulties for the model training. So in this preliminary study, we choose to apply this constraint in our primary dataset $Data_{exp}$ and perform an additional experiment later in Section \ref{rq4} to investigate the performance of our models without this constraint.

In addition, we notice that the NLTK tokenizer\footnote{\url{http://www.nltk.org/api/nltk.tokenize.html}} can not separate special tokens in code snippets well, which leads to an extensive vocabulary and exacerbates the out-of-vocabulary issue.
So we choose a simple tokenizing algorithm to tackle this problem. Specifically, there are three kinds of printable characters in the ASCII charset, including the digits (\emph{0 to 9}), letters (\emph{A/a to Z/z}), and punctuation symbols.  We first put a white space on both sides of punctuation symbols in a string during tokenization and then split the string into tokens by white spaces. This way, we get a smaller vocabulary but longer sequences in return. Handling very long sequences is still an open problem in the field of deep learning \cite{Dai2019TransformerXLAL, Beltagy2020LongformerTL, Zaheer2020BigBT}. In this preliminary study, we choose to filter out 5\% of the question posts whose body length exceeds 1,000, or the title length exceeds 25.

As for data partitioning, we sort the questions in chronological order and choose the latest samples for testing and the rest for training. Because we think it is more applicable to the real-world application if the models take past questions for training and new ones for testing. Besides, we believe our time-wise partitioning will help relieve the target leakage problem caused by the homogeneous questions between the train and test sets. We show the statistics of our processed dataset in Table \ref{tbl1}.
\begin{table}[pos=h]
	\caption{The partition size of $Data_{exp}$}\label{tbl1}
	\begin{tabular}{@{\ \ \ }cccc@{\ \ \ }}
		\toprule
		Language & Train & Validation & Test\\ \midrule
		Java & 57,118 & 2,000 & 2,000 \\
		Python & 60,458 & 2,000 & 2,000 \\
		JS & 53,708 & 2,000 & 2,000 \\
		PHP & 26,535 & 1,000 & 1,000\\ \midrule[0.2pt]
		Total & 197,819 & 7,000 & 7,000\\
		\bottomrule
	\end{tabular}
\end{table}

\subsection{The CCBERT Model}
\begin{figure}
	\centering
	\includegraphics[scale=.15]{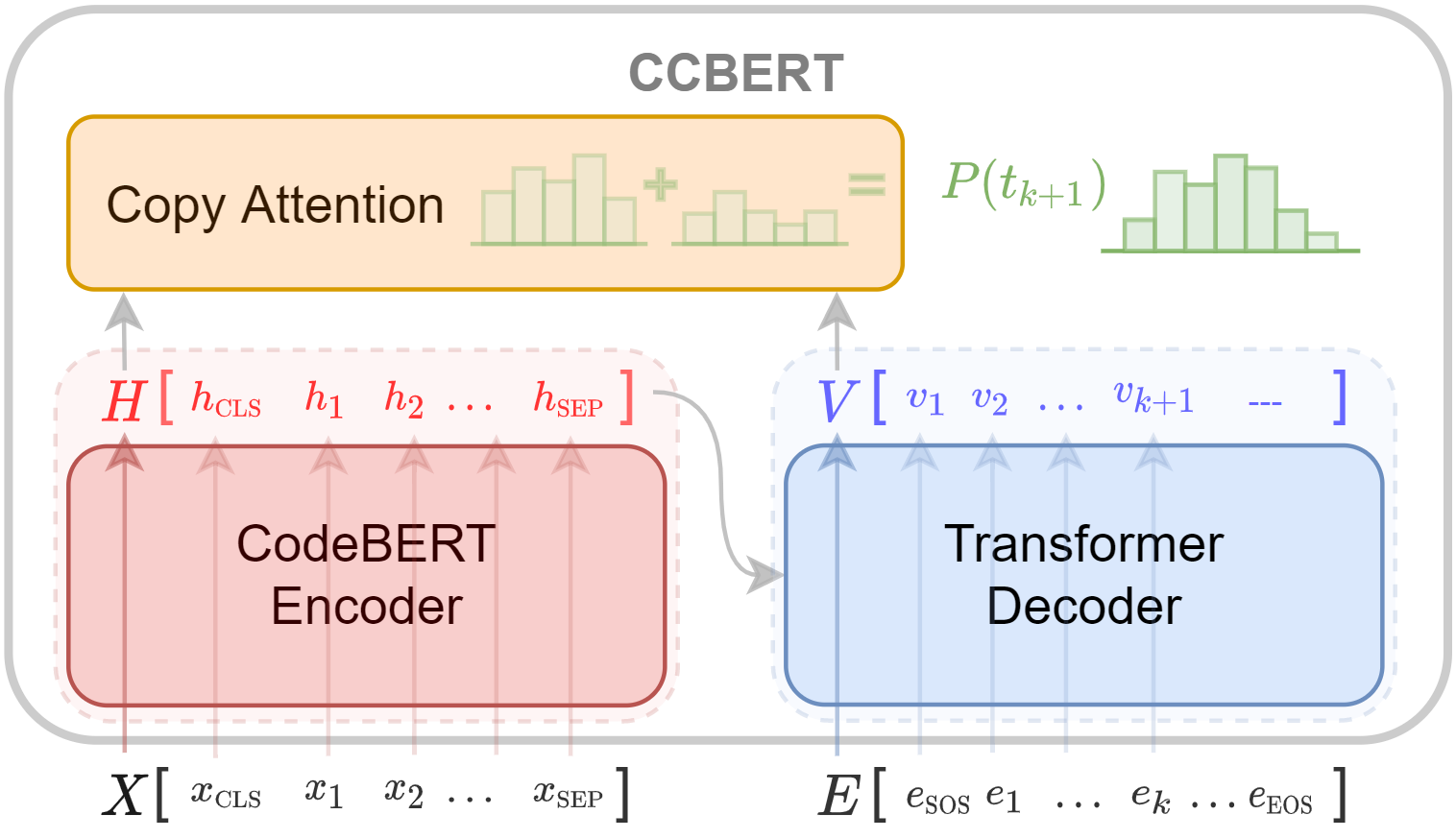}
	\caption{The detailed structure of CCBERT at the $(k+1)th$ decoding step}
	\label{FIG:Fig. 5}
\end{figure}

We propose CCBERT, a novel model combining the pre-trained CodeBERT model and the copy mechanism. It is an attentional encoder-decoder system, which can be trained and used in an end-to-end manner. Our model architecture is illustrated in Fig. \ref{FIG:Fig. 5}. Specifically, we apply the CodeBERT model and Transformer-decoder layers in their original form and put a specialized copy attention layer above the encoder and decoder. 
Formally, given a token sequence $X=[x_1,x_2,...x_n]$ of a question body and a token sequence $T=[t_1,t_2,...,t_l]$ of its corresponding title sampled from our dataset, CCBERT learns to generate $T$ based on $X$.

\subsubsection{CodeBERT Encoder}
Unlike the general models used for summarization \cite{Gehrmann2018BottomUpAS, Liu2019TextSW, See2017GetTT}, our model needs to understand both Natural Language (NL) and Programming Language (PL), which is determined by the characteristics of our dataset. Recently, Feng et al. \cite{Feng2020CodeBERTAP} introduced CodeBERT, which was pre-trained on a vast scale dataset extracted from Github repositories containing source code and code comments. This way, CodeBERT can capture the semantic relationship between NL and PL, and produce vector representations that support downstream tasks, such as defect prediction \cite{Pan2021AnES, ZHAO2021106652, Zhao2021PredictingCF, Zhao2020SimplifiedDF, zhao2022graph4web}, program repair \cite{Mashhadi2021ApplyingCF}, etc. 

We use the pre-trained CodeBERT as our encoder. It is a stack of multiple Transformer-encoder layers which mainly performs bidirectional self-attention operations. Formally, given a question body containing text descriptions and code snippets, we first turn it into a sequence of tokens with a byte pair encoding tokenizer built in the CodeBERT model. Then, we surround the token sequence with two special tokens\footnote{The \emph{SEP} token marks the end of a sentence. The \emph{CLS} token is put in front of the input sequence and specially used for sentence classification.} to be consistent with the data format used during pre-training and get the final input sequence $X$,
\begin{equation}
    X=\left[x_{CLS},x_1,x_2,\ldots,x_n,x_{SEP}\right].
\end{equation}
After that, we feed $X$ to the encoder and get a matrix $H$ that consists of the encoded vectors of all input tokens
\begin{equation}
	H=\mathrm{ENCODER}(X),
\end{equation}
where $H=[h_{CLS},h_1,h_2,...h_n,h_{SEP}]$ and each vector $h_i$ is a hidden representation of the semantic relationship of a token against others.

\subsubsection{Transformer Decoder}
After encoding the input question body, we need the decoder to generate the hidden representation of each token in the predicted question title. Since the nature of the CodeBERT encoder is Transformer-encoder layers, we stack several layers of vanilla Transformer-decoder \cite{Vaswani2017AttentionIA} as our decoder.

Formally, suppose we have generated the first $k$ tokens (i.e., $Y=\left[y_{SOS},y_1,y_2,\ldots,y_k\right]$)\footnote{$SOS$ means the start of a sequence. } of the predicted title, and now are going to generate the $\left(k+1\right)th$ token (i.e, at the $(k+1)th$ decoding step).
We first use the same embedding layer of the encoder to turn the input sequence $Y$ into a matrix $E$ containing the embedding vectors of tokens (i.e, $E=\left[e_{SOS},e_1,e_2,\ldots,e_k\right]$). The input for the decoder is two-fold, one is the hidden vectors $H$ provided by the encoder, the other is the embedding $E$ of generated sequence. We feed $H$ and $E$ to the decoder and get a matrix $V$ containing the hidden representations of $k+1$ predicted tokens
\begin{equation}
	V=\mathrm{DECODER}(H, E),
\end{equation}
where $V=[v_1,v_2,...,v_{k+1}]$. We take $v_{k+1}$ as the hidden representation of the $\left(k+1\right)th$ predicted token.

\subsubsection{Copy Attention Layer}
Usually, we can have several linear layers above the decoder to map the hidden representation $v_{k+1}$ to its most likely token in the vocabulary. However, according to the statistics, question titles have a high overlap with the body content. Besides, we should also pay attention to some essential but rare tokens, such as variable names, class libraries, application frameworks, etc. In this case, we incorporate the copy mechanism to facilitate our model to copy tokens directly from the body content when generating titles. The copy mechanism was first introduced in the pointer-generator network \cite{See2017GetTT}, which was originally applied to the Recurrent Neural Networks (RNNs). In our work, we implement a specialized copy attention layer to adapt the copy mechanism to our Transformer-based model. 

Formally, when generating the $\left(k+1\right)th$ token in the predicted title, we first need to calculate the attention vector $a_{k+1}$ with the encoder hidden state $H$, the embedding vector of the generated token $e_k$, and the hidden representation $v_{k+1}$ of the $(k+1)th$ predicted token,
\begin{equation}
    a_{k+1}=Attention(H,e_k,v_{k+1}).
\end{equation}
Then, we use the attention vector $a_{k+1}$ and the encoder hidden state matrix $H$ to get a single vector $context_{k+1}$ as the overall "context" of the input sequence,
\begin{equation}
	context_{k+1}^\mathsf{T}=a_{k+1}^\mathsf{T}\cdot H,
\end{equation}
where $\mathsf{T}$ represents the transpose symbol. After that, with the input context $context_{k+1}$ and the current decoder state $v_{k+1}$, we can get the probability distribution $P_{vocab}$ of each token in the vocabulary to be chosen as the $(k+1)th$ predicted token,
\begin{equation}
	P_{vocab}=LinearSoftmax(context_{k+1},v_{k+1}),
\end{equation}
where $LinearSoftmax$ is a linear neural network with the Softmax output layer.

Usually, we can choose the token with the highest probability in $P_{vocab}$ as the predicted token. But to incorporate the copy mechanism, we have to calculate an additional probability $p_{copy}$  as a soft switch to choose between generating a word from the vocabulary by sampling from $P_{vocab}$, or copying a word from the input sequence by sampling from the attention distribution $a_{k+1}$,
\begin{equation}
	p_{copy}=LinearSigmoid(context_{k+1},v_{k+1},e_k),
\end{equation}
where $LinearSigmoid$ is a linear layer with the Sigmoid activation function.
%\begin{figure}
%	\centering
%	\includegraphics[scale=.23]{figs/figure-5.png}
%	\caption{The Copy Attention Layer}
%	\label{FIG:Fig. 7}
%\end{figure}
Finally, we can get the revised probability distribution $P(t_{k+1})$ of choosing the $\left(k+1\right)th$ token,
\begin{equation}
	P(t_{k+1})=p_{copy}\sum_{i:x_i=t_{k+1}}^{n}a_{(k+1)_i}+(1-p_{copy})P_{vocab}(t_{k+1}).
\end{equation}
We generate each token recursively and stop when the $EOS$\footnote{$EOS$ means the end of a sequence} token comes up. The overall trainable parameters $\theta$ include those of the stacked Transformer-encoder layers in CodeBERT, the stacked Transformer-decoder layers in our decoder, and the linear neural networks in our copy attention layer. The training loss is the negative log-likelihood of each token in the target sequence, which we use to update $\theta$ through backpropagation to maximize the
likelihood between the generated titles and the original ones in our dataset during training.

\section{Experimental Setup}
\label{experimental setup}

This section illustrates the baseline models, the evaluation metrics, and the hyperparameter settings for our CCBERT model.

\subsection{Comparisons}
\label{baselines}
To demonstrate how competitive CCBERT is, we choose several state-of-the-art models as baselines, which have been widely studied in the field of natural language processing. We briefly introduce the general ideas of these models in the following.
\begin{enumerate}[(1)]
	\item \textbf{TF-IDF} \hspace{3mm} This method is a classic full text searching algorithm, its name stands for "Term Frequency (TF)\ $\times$ Inverse Document Frequency (IDF)".
	TF-IDF is a weighting algorithm for a bag-of-words language model. Specifically, the "bag" contains a list of unique terms sourced from a given corpus. A paragraph can be turned into a vector by counting its in-bag terms' frequency (TF). Because the probability of a term's occurrence is often in inverse proportion to its importance, one can use the term frequency of appearing in all documents ($\mathrm{IDF}^{-1}$) to divide TF and get the revised weight of each term. This way, we can calculate the distance between paragraphs in the vector space. In our experiment, we use Lucene\footnote{Apache Lucene computes the similarity using TF-IDF by default.} to find the most similar question in the train set given a question body.
	
	\item \textbf{BiLSTM} \hspace{3mm} Long Short Term Memory networks (LSTMs) are a special kind of RNNs, with an additional cell state and three carefully designed "gates" to alleviate the problem of long-term dependencies. 
	The idea of Bidirectional LSTMs (BiLSTMs) is to duplicate the first recurrent layer in the network and then provide the input sequence to the first layer and a reversed copy of the input to the second. This way, all available information in the past and future of a specific processing step can be considered during training. We stack two BiLSTM layers as the encoder and two LSTM layers as the decoder, along with the attention mechanism introduced by Bahdanau et al. \cite{Bahdanau2015NeuralMT} to build a model as our baseline, which we refer to as BiLSTM.
	
	\item \textbf{BiLSTM-CC} \hspace{2mm} This was the method used by Gao et al. \cite{gao2020generating} to generate question titles from code snippets. It shares the same structure as the BiLSTM model mentioned above, except to assemble another two non-trivial mechanisms. One is the copy mechanism we have illustrated above; the other is the coverage mechanism. Tu et al. \cite{Tu2016ModelingCF} first introduced the "coverage" vector that keeps track of the attention history and further facilitates the attention calculation so that a neural machine translation system would consider more about untranslated words. Gao et al. \cite{gao2020generating} took advantage of the coverage penalty to suppress meaningless repetitions during generation. In our experiment, we build the BiLSTM and BiLSTM-CC models with OpenNMT,\footnote{\url{https://opennmt.net}} which is a well-acknowledged framework to build sequence-to-sequence models.
	
	\item \textbf{BART} \hspace{3mm} Lewis et al. \cite{Lewis2020BARTDS} proposed the \underline{BART} model to bridge the gap between pre-trained \underline{B}idirectional encoder (i.e. BERT \cite{Devlin2019BERTPO}) and pre-trained \underline{A}uto-\underline{R}egressive \underline{T}ransformer (i.e. GPT \cite{Radford2019LanguageMA}), which are good at comprehension and generation tasks respectively. BART is pre-trained under the supervision of several denoising objectives, where input text is corrupted by a stochastic noising function and the model is demanded to reconstruct the original text. BART is particularly effective when fine-tuned for neural machine translation and abstractive text summarization tasks, such as WMT, CNN/DailyMail, XSum etc. We use the open-source code and pre-trained parameters\footnote{\url{https://huggingface.co/facebook/bart-base}} for BART to validate its performance on our dataset. 
\end{enumerate}

In addition to the above baselines, we implement an oracle method to show the best performance of an extractive model.

\begin{enumerate}[(5)]
	\item \textbf{Oracle} \hspace{3mm} The idea of extractive summarization, which is to select primary sentences that best match the target summary, inspires us to explore the possibility of making up a title only using tokens that appear in the question body. 
	However, there are millions of permutations of a title containing tens of tokens, which is more complicated than selecting and arranging sentences. In addition, tokens arranged in the correct order do not necessarily make sense. So, instead of building another baseline model, we remove the tokens in a question title if they are not in the question body and keep the rest as the "generated" title to simulate the best performance of an extractive model. Considering our objective is to maximize the BLEU and ROUGE score, we follow the work of Liu et al. \cite{Liu2019TextSW} and implement another method based on beam search (with 20 beam width) to find the permutation that performs the best on these two metrics. 
	It turns out that the second method does no better than the first one on both metrics due to the limited searching space. Therefore, we use the simple method mentioned above as the oracle method indicating the best possible result from a model.
	
\end{enumerate}

\subsection{Automated Evaluation Metrics}
Since the nature of our task is a sequence generation problem, where BLEU and ROUGE are the most commonly used metrics, we choose both to measure the precision- and recall-oriented similarity between the generated titles and the original ones.

\subsubsection{BLEUS-4}
The \underline{B}i-\underline{L}ingual \underline{E}valuation \underline{U}nderstudy (\underline{BLEU}) method was first introduced by Papineni et al. \cite{Papineni2002BleuAM} to measure the performance of a translation system. Given the candidate translations and reference sentences, the first step in this method is to compute the $ngram$ precision, 
\begin{equation}
	p_n=\frac{\sum\limits_{C\in \{candidates\}}\sum\limits_{ngram\in C}Count_{clip}(ngram)}{\sum\limits_{C\in \{candidates\}}\sum\limits_{ngram\in C}Count(ngram)},
\end{equation}
\begin{equation}
	Count_{clip}=min(Count, Max\_ref\_Count),
\end{equation}
where $ngram$ denotes the candidate ngrams, $Count_{clip}$ clips the total $Count$ of each candidate ngram by the maximum number of overlap between $ngram$ in the candidate and the references $Max\_ref\_Count$, to avoid overgenerating "reasonable" words. In brief, the numerator of $p_n$ counts the number of candidate ngrams that appear in references, and the denominator counts all the candidate ngrams.

The next step is to compute a brevity penalty, which is to adapt the candidate translation to match the reference translation in length,
\begin{equation}
	{BP}=
	\begin{cases}
		1, &{\rm{if}} \ l_c>l_r\\
		e^{(1-l_r/l_c)}, &{\rm{if}} \ l_c\le l_r
	\end{cases},
\end{equation}
where $l_c$ is the length of a candidate translation and $l_r$ is the length of the reference corpus. Then, we can get the BLEU score
\begin{equation}
	\mathrm{BLEU}=BP\cdot exp\Bigg(\sum\limits_{n=1}^N\frac{1}{N}logp_n\Bigg).
\end{equation}
In our experiment, we choose $N=4$ to have the BLEU-4 score. Besides, we apply a smoothing method introduced by Lin et al. \cite{Lin2004ORANGEAM} to add one to the $ngram$ hit count and total $ngram$ count for $n>1$. This way, candidate translations with less than \textit{n} words can still get a positive score. We refer to the smoothed method as BLEUS-4 and use its implementation of NLTK\footnote{\url{http://www.nltk.org/_modules/nltk/translate/bleu_score.html}} in our experiment.

\subsubsection{ROUGE}
The \underline{R}ecall-\underline{O}riented \underline{U}nderstudy for \underline{G}isting \underline{E}valuation (\underline{ROUGE}) was introduced by Lin et al. \cite{Lin2004ROUGEAP} to measure the quality of machine-generated summaries. It consists of several measures including ROUGE-N and ROUGE-L, which will be used in our experiment. On complementary of BLEU's bias on $ngram$ precision, ROUGE-N focuses on the $ngram$ recall, which is calculated as
\begin{equation}
	\operatorname{ROUGE-N}=\frac{\sum\limits_{S\in \{References\}}\sum\limits_{ngram\in S}Count_{m}(ngram)}{\sum\limits_{S\in \{References\}}\sum\limits_{ngram\in S}Count(ngram)},
\end{equation}
where $ngram$ denotes the reference ngrams, $Count_{m}$ is the maximum number of overlap between $ngram$ in a candidate summary and the references. In brief, the numerator of ROUGE-N is to count the number of overlap ngrams between candidates and references, and the denominator is to count all the reference ngrams.

ROUGE-L takes advantage of both the Longest Common Subsequence (LCS) and the F-measure to estimate the similarity between two summaries: the candidate summary $S_{can}$ of length $l_a$ and the reference summary $S_{ref}$ of length $l_e$. The calculation is as follows:
\begin{equation}
	Recall_{lcs}=\frac{LCS(S_{ref},S_{can})}{l_{e}},
\end{equation}
\begin{equation}
	Precision_{lcs}=\frac{LCS(S_{ref},S_{can})}{l_{a}},
\end{equation}
\begin{equation}
	\operatorname{ROUGE-L}=\frac{Recall_{lcs}Precision_{lcs}}{Recall_{lcs}+Precision_{lcs}},
\end{equation}
In the end, we choose ROUGE-1, ROUGE-2, and ROUGE-L implemented by an open source library\footnote{\url{https://pypi.org/project/rouge}} for the evaluation metrics.

\subsection{Model Settings}
We implement our encoder with the pre-trained parameters\footnote{\url{https://huggingface.co/microsoft/codebert-base}} of CodeBERT-base and keep its initial settings, where the vocabulary size is 50265, the hidden size is 768, the dropout probability is 0.1, and the Transformer layer number is 12.
Accordingly, we build a 12-layer decoder with randomly initialized parameters. Optimization is performed using the adaptive moment estimation (Adam) algorithm with $\beta_1=0.9$, $\beta_2=0.999$, $\epsilon={10}^{-8}$ and $lr=5\times{10}^{-5}$. We also apply a linear warm-up strategy to gradually increase the learning rate in the first 10\% training steps. Four NVIDIA GeForce RTX 2080 Ti GPUs are used to train our model, where the training epoch is ten and batch size is 32. During decoding, we set the beam size to ten. We adjust all the hyperparameters to the validation set and report the evaluation results on the test set.

\section{Results and Analysis}
\label{result}
\label{results and analysis}

\begin{table*}[p]
	\centering
	\caption{The evaluation results of models trained on the joint dataset of four programming languages. All the score numbers are averages over the tested posts of different languages.}\label{tbrq1}
	\resizebox{0.65\textwidth}{!}{
	\begin{tabular}{@{\ \ }cccccc@{\ \ }}
		\toprule
		Model                  & Language     & BLEUS-4 & ROUGE-1 & ROUGE-2 & ROUGE-L \\ \midrule[0.6pt]
		\multirow{4}{*}{Oracle}
		& \scriptsize Java    &  54.58 & 83.76 & 65.13 & 83.31 \\
		& \scriptsize Python    &  51.43 & 82.40 & 62.55 & 81.81 \\
		& \scriptsize JS      &  53.10 & 83.02 & 63.68 & 82.55  \\
		& \scriptsize PHP    & 54.22 & 83.65 & 65.19 & 83.21 \\ \midrule[0.6pt]
		\multirow{4}{*}{TF-IDF}
		& \scriptsize Java    & 9.79 & 19.91 & 4.44 & 19.17  \\
		& \scriptsize Python    & 10.26 & 21.88 & 5.28 & 21.01  \\
		& \scriptsize JS    &  10.10 & 20.51 & 4.93 & 19.76  \\
		& \scriptsize PHP    & 10.49 & 21.24 & 5.15 & 20.30 \\ \midrule[0.2pt]
		\multirow{4}{*}{$\mathrm{BiLSTM_{joint}}$}
		& \scriptsize Java    & 17.04 & 36.74 & 15.35 & 36.17  \\
		& \scriptsize Python    & 17.71 & 39.89 & 16.86 & 39.04  \\
		& \scriptsize JS &  18.06 & 38.79 & 16.56 & 38.09  \\
		& \scriptsize PHP    & 18.66 & 39.97 & 17.99 & 38.92 \\ \midrule[0.2pt]
		\multirow{4}{*}{BiLSTM-$\mathrm{CC_{joint}}$}
		& \scriptsize Java    & 19.73 & 41.10 & 19.54 & 40.04  \\
		& \scriptsize Python    & 19.74 & 42.67 & 19.97 & 41.72  \\
		& \scriptsize JS &  20.59 & 42.62 & 20.36 & 41.59  \\
		& \scriptsize PHP    & 20.56 & 43.01 & 20.92 & 41.73 \\ \midrule[0.2pt]
		\multirow{4}{*}{$\mathrm{BART_{joint}}$}
		& \scriptsize Java    & 20.80 & 44.21 & 21.12 & 42.42  \\
		& \scriptsize Python    & 21.01 & 45.69 & 22.44 & 44.28  \\
		& \scriptsize JS &  21.54 & 45.65 & 22.29 & 43.81  \\
		& \scriptsize PHP    & 22.28 & 46.94 & 23.47 & 45.05 \\ \midrule[0.2pt]
		\multirow{4}{*}{$\mathrm{CCBERT_{joint}}$}
		& \scriptsize Java    & \textbf{21.16} & \textbf{44.26} & \textbf{21.58} & \textbf{42.92}  \\
		& \scriptsize Python    & \textbf{22.40} & \textbf{46.88} & \textbf{22.89} & \textbf{44.92}  \\
		& \scriptsize JS	& \textbf{22.18} & \textbf{45.72} & \textbf{22.40} & \textbf{44.15}  \\
		& \scriptsize PHP    & \textbf{22.65} & \textbf{47.03} & \textbf{23.50} & \textbf{45.15}  \\ \bottomrule
	\end{tabular}
	}
\end{table*}

\begin{table*}[p]
	\centering
	\caption{The evaluation results of models trained on the separate datasets of four programming languages. All the score numbers are averages over the tested posts of different languages. The Oracle and TF-IDF models are not affected by separated training.}
	\label{tbrq1_2}
	\resizebox{0.65\textwidth}{!}{
	\begin{tabular}{@{\ \ }cccccc@{\ \ }}
		\toprule
		Model                  & Language     & BLEUS-4 & ROUGE-1 & ROUGE-2 & ROUGE-L \\ \midrule[0.6pt]
		\multirow{4}{*}{$\mathrm{BiLSTM_{sep}}$}
		& \scriptsize Java    & 14.59  &  32.13  &   11.95 &   31.86  \\
		& \scriptsize Python    & 15.93   &   37.15	&   14.24 &  36.55  \\ %
		& \scriptsize JS &  14.91 & 33.03 & 11.63 & 32.50  \\
		& \scriptsize PHP    & 13.03 & 28.41 & 08.73 & 27.62 \\ \midrule[0.2pt]
		\multirow{4}{*}{BiLSTM-$\mathrm{CC_{sep}}$}
		& \scriptsize Java    & 18.22  &  38.89  &   18.09 &   38.21  \\
		& \scriptsize Python    & 18.84  &   41.49 &   18.98 &  40.56  \\
		& \scriptsize JS &  19.35 & 40.92 & 18.28 & 40.15  \\
		& \scriptsize PHP    & 19.08 & 40.83 & 18.75 & 39.72 \\ \midrule[0.2pt]
		\multirow{4}{*}{$\mathrm{BART_{sep}}$}
		& \scriptsize Java    & 19.32  &  42.52  &   19.93 &   41.66  \\
		& \scriptsize Python    & 20.43  &  44.93  &   21.77 &  43.97  \\
		& \scriptsize JS &  20.19 & 43.54 & 20.56 & 42.70  \\
		& \scriptsize PHP    & 19.61 & 43.50 & 20.79 & 42.65 \\ \midrule[0.2pt]
		\multirow{4}{*}{$\mathrm{CCBERT_{sep}}$}
		& \scriptsize Java    & \textbf{20.90}  &  \textbf{43.06}  &   \textbf{21.15} &   \textbf{41.76}  \\
		& \scriptsize Python    & \textbf{22.02}  &  \textbf{46.69}  &  \textbf{22.55}  &  \textbf{44.86}  \\
		& \scriptsize JS    & \textbf{21.24} & \textbf{44.55} & \textbf{21.05} & \textbf{42.80}  \\
		& \scriptsize PHP    & \textbf{21.93} & \textbf{45.60} & \textbf{22.35} & \textbf{43.87}  \\ \bottomrule
	\end{tabular}
	}
\end{table*}

\begin{table*}[p]
	\centering
	\caption{The examples of our testing questions and automatically generated titles. Specifically, the green color marks the tokens appearing in original titles, the orange-red color marks the wrong focus, and the gray color marks the code snippets. The models with a $code$ subscript in their names are trained on the code-only dataset. }\label{tbrq1example}
	\resizebox{0.90\textwidth}{!}{
	\begin{tabular}{@{\ \ }l|l@{\ }}
		\toprule
		Question Body  & Titles \\ \midrule[0.6pt]
		\multirow{12}{8.2cm}{
			I need to create this shape. I understand how to create simple shapes such as a cube, but I don't understand at all how to \textcolor{LimeGreen}{create such} a \textcolor{LimeGreen}{shape}. \textcolor{OrangeRed}{How to get the right points for these arrays}? Please, help\\
			\textcolor{gray}{
				TriangleMesh mesh = new \textcolor{LimeGreen}{TriangleMesh}();\\
				\hspace{4mm}mesh.getPoints().addAll(...\\
				\hspace{4mm}mesh.getTexCoords().addAll(...\\
				\hspace{4mm}//which points should be here\\
				\hspace{4mm}mesh.getFaces().addAll(...\\
				\hspace{4mm}//which points should be here\\
				\hspace{4mm}return mesh;\\
			}
		}
		& \underline{\textbf{\scriptsize{Origin}}: \href{https://stackoverflow.com/questions/61231437/how-to-create-such-shape-using-javafx-trianglemesh}{how to create such shape using javafx trianglemesh}} \\
		& \textbf{\scriptsize{Oracle}}: how to create such shape trianglemesh \\
		& \textbf{\scriptsize{TF-IDF}}: how update the value in json file using java jackson \\
		& \textbf{\scriptsize{BiLSTM}}: \textcolor{OrangeRed}{how to get the right points for arrays} \\ %
		& \textbf{\scriptsize{BiLSTM-CC}}: how to \textcolor{LimeGreen}{create} such a \textcolor{LimeGreen}{shape} in java \\
		& \textbf{\scriptsize{BART}}: how to \textcolor{LimeGreen}{create} such a \textcolor{LimeGreen}{shape} \\
		& \textbf{\scriptsize{CCBERT}}: how to \textcolor{LimeGreen}{create} this \textcolor{LimeGreen}{shape} using \textcolor{LimeGreen}{trianglemesh} \\ 
		& \\
		& \textbf{\scriptsize{BiLSTM-CC$_{code}$}}: how to get points of a mesh \\
		& \textbf{\scriptsize{CCBERT$_{code}$}}: how to add points to a \textcolor{LimeGreen}{trianglemesh} \\
		& \\ \midrule[0.4pt]

		\multirow{15}{8.2cm}{
			I am a beginner in mobile application building. I tried to put \textcolor{LimeGreen}{insert data} function in my \textcolor{OrangeRed}{android studio} but those insert function doesn't work and the \textcolor{LimeGreen}{input data} can't be inserted...\\
			\ I put code in MainActivity.java and DatabaseHelper.java. It doesn't give error report but when run the emulator and input data, my input can be inserted to \textcolor{LimeGreen}{sqlite} database.\\
			\textcolor{gray}{
				//oncreateMainActivity\\
				super.onCreate(savedInstanceState...\\
				\hspace{4mm}myDb = new DatabaseHelper(...\\
				\hspace{4mm}submit2.setOnClickListener(...\\
				//DatabaseHelper.java\\
				public boolean insertData(String...\\
				\hspace{4mm}SQLiteDatabase db = this.getWritableDatabase(...\\
				\hspace{4mm}long result = db.insert(TABLE\_NAME...\\
			}
		}
		& \underline{\textbf{\scriptsize{Origin}}: \href{https://stackoverflow.com/questions/55550279/how-to-insert-data-to-sqlite-through-user-input}{how to insert data to sqlite through user input}} \\
		& \textbf{\scriptsize{Oracle}}: to insert data to sqlite input  \\
		& \textbf{\scriptsize{TF-IDF}}: listview not show items stored in \textcolor{LimeGreen}{sqlite} database \\
		& \textbf{\scriptsize{BiLSTM}}: how to put function in my \textcolor{OrangeRed}{android studio} \\
		& \textbf{\scriptsize{BiLSTM-CC}}: how to \textcolor{LimeGreen}{insert data} function in \textcolor{OrangeRed}{android studio} \\
		& \textbf{\scriptsize{BART}}: how to \textcolor{LimeGreen}{insert data} in \textcolor{OrangeRed}{android studio} \\
		& \textbf{\scriptsize{CCBERT}}: how to insert \textcolor{LimeGreen}{input data} to \textcolor{LimeGreen}{sqlite} database \\ 
		& \\
		& \textbf{\scriptsize{BiLSTM-CC$_{code}$}}: how to add data to an activity in android \\
		& \textbf{\scriptsize{CCBERT$_{code}$}}: how to get data from database in android \\
		& \\ & \\ & \\ & \\ & \\ \midrule[0.4pt]
		
		\multirow{12}{8.2cm}{
			I have a simple web application where different users can log into it...send email of it's content to \textcolor{OrangeRed}{an outsider} like third party....With all this, I am using \textcolor{OrangeRed}{Java Mail API} to make it work and after hitting the send button,it sends directly to the recipient...Now, I want to modify this by doing this email feature as a \textcolor{LimeGreen}{service}...the content and info filled in will be stored in a table in \textcolor{OrangeRed}{MYSQL} and...\\
			\textcolor{gray}{
    			public void sendEmail(String ... \{\\
                \hspace{4mm}Properties props = new Properties();\\
                \hspace{4mm}props.put("mail.smtp.host", host); //SMTP...\\
                \hspace{4mm}Authenticator auth = new Authenticator()...\\
            }
            Can this be done in the way...how to make it work?\\
		}
		& \underline{\textbf{\scriptsize{Origin}}: \href{https://stackoverflow.com/questions/55585141/java-how-to-use-services-for-sending-email}{java - how to use services for sending email}} \\
		& \textbf{\scriptsize{Oracle}}: java how to sending email \\
		& \textbf{\scriptsize{TF-IDF}}: gae send email from gmail account \\
		& \textbf{\scriptsize{BiLSTM}}: how can i send email to  \textcolor{OrangeRed}{an outsider} \\
		& \textbf{\scriptsize{BiLSTM-CC}}: how to send email  \textcolor{OrangeRed}{from database} in java \\
		& \textbf{\scriptsize{BART}}: how to send email using \textcolor{OrangeRed}{java mail api} \\
		& \textbf{\scriptsize{CCBERT}}: how to send email using \textcolor{LimeGreen}{services} in java \\ 
		& \\ 
		& \textbf{\scriptsize{BiLSTM-CC$_{code}$}}: how to set the header of a mail in a mail \\
		& \textbf{\scriptsize{CCBERT$_{code}$}}: how to send email using java mail \\
		& \\ & \\ \midrule[0.4pt]
		
		\multirow{11}{8.2cm}{
			I want to work with \textcolor{OrangeRed}{crypto-stock} data described here in my \textcolor{OrangeRed}{spring boot} application. The RESTTemplate uses Gson for \textcolor{LimeGreen}{deserialization}. Response data looks like: \\
			\textcolor{gray}{
				\{"IOST": \{"EUR": 0.01147,… \\
			}
			I have already...problem is that this comes as a single object with \textcolor{LimeGreen}{key-value} pairs insted of as an array. The result should be a list of following objects: \\
			\textcolor{gray}{
				public class Symbol \{ \\
				\hspace{4mm}private Long id; \\
				\hspace{4mm}private String symbol...\\
			}
			Any idea how this can be accomplished this? \\
		}
		& \underline{\textbf{\scriptsize{Origin}}: \href{https://stackoverflow.com/questions/55704173/how-to-deserialize-a-key-value-map-to-a-list}{how to deserialize a key-value map to a list}} \\
		& \textbf{\scriptsize{Oracle}}: how to a key-value to a list  \\
		& \textbf{\scriptsize{TF-IDF}}: bigdecimal not keeping actual value when returned \\
		& \textbf{\scriptsize{BiLSTM}}:  how to work with \textcolor{OrangeRed}{crypto-stock data} \\
		& \textbf{\scriptsize{BiLSTM-CC}}: how to parse json data in \textcolor{OrangeRed}{spring boot} \\
		& \textbf{\scriptsize{BART}}: how to \textcolor{LimeGreen}{deserialize} \textcolor{OrangeRed}{crypto-stock} data \\
		& \textbf{\scriptsize{CCBERT}}: how to \textcolor{LimeGreen}{deserialize} \textcolor{LimeGreen}{key-value} pairs \\ 
		& \\
		& \textbf{\scriptsize{BiLSTM-CC$_{code}$}}: how to convert json object to java object \\
		& \textbf{\scriptsize{CCBERT$_{code}$}}: how to \textcolor{LimeGreen}{deserialize} a json object in java \\
		& \\
		\bottomrule
	\end{tabular}
}
\end{table*}

\begin{table*}
	\centering
	\caption{
		The performance of CCBERT and BiLSTM-CC on the code-only dataset. All the score numbers are averages over the tested posts of different languages.
	}\label{tbl4}
	\resizebox{0.65\textwidth}{!}{
	\begin{tabular}{@{\ }cccccc@{\ }}
		\toprule
		Model                  & Language      & BLEUS-4 & ROUGE-1 & ROUGE-2 & ROUGE-L \\ \midrule[0.6pt]
		\multirow{4}{*}{BiSLTM-$\mathrm{CC_{code}}$}
        & \scriptsize Java & 11.78 & 25.04 & 7.15 & 25.50 \\
        & \scriptsize Python & 13.38 & 30.54 & 9.78 & 30.48 \\
        & \scriptsize JS & 13.02 & 28.00 & 8.35 & 28.18 \\
        & \scriptsize PHP & 13.13 & 29.25 & 9.09 & 28.63 \\ \midrule[0.4pt]
		\multirow{4}{*}{$\mathrm{CCBERT_{code}}$}
        & \scriptsize Java & 12.84 & 28.73 & 9.62 & 28.58 \\
        & \scriptsize Python & 14.03 & 33.35 & 11.67 & 32.67 \\
        & \scriptsize JS & 13.67 & 30.57 & 10.36 & 30.20 \\
        & \scriptsize PHP & 14.32 & 32.80 & 11.90 & 31.88 \\ \bottomrule
	\end{tabular}
	}
\end{table*}

In this section, we demonstrate the effectiveness of our model by conducting experiments to answer the following Research Questions (RQs):
\begin{enumerate}[RQ-1]
	\item Does our CCBERT model outperform the baseline models?
	\item What is the advantage of using the bi-modal information of the entire question body?
	\item How effective is our CCBERT model under low-resource circumstances?
	\item How much influence does interrogative constraint have on model training?
	\item How effective is our CCBERT model under human evaluation?
\end{enumerate}

\subsection{RQ-1: Does our CCBERT model outperform the baseline models?}
\label{rq1}
\noindent \textbf{Method: }In order to investigate the superiority of our model, we compare it to the baselines mentioned in Section \ref{baselines}. We also apply two training strategies to the deep learning models. One is to train on the $Data_{exp}$ jointly with all the questions. The other is to train on separated smaller subsets of questions concerning different programming languages. Both training strategies share the same validation and test sets to compare the performance.
Table \ref{tbrq1} and Table \ref{tbrq1_2} show the performance of all models on four evaluation metrics. In addition, we present four test examples in Table \ref{tbrq1example} to make intuitive comparisons.

\noindent \textbf{Results: }From Table \ref{tbrq1}, Table \ref{tbrq1_2}, and Table \ref{tbrq1example}, we have the following findings: 

(1) The performance rankings are the same on both training strategies, where CCBERT outperforms all the baselines ranging from the retrieval-based model (TF-IDF) to the large-scale pre-trained model (BART). Moreover, all the models trained on the joint dataset perform better than those trained on separated subsets, attributing to the increased amount of training samples and the similar writing pattern shared by high-quality questions involving different programming languages.
We have also noticed that Java questions are more difficult for all the models, which is similar to the results reported by Gao et al. \cite{gao2020generating}. This is partly because Java questions have a larger vocabulary than others, and models are more likely to encounter rare tokens.

(2) TF-IDF has the worst performance among all the baseline models and can barely compare with other baselines. This is not surprising because questions containing duplicated content in Stack Overflow have a high possibility of being closed, let alone only a small number of questions available in our train set. Besides, the nature of TF-IDF is a bag-of-word model, which does not take into account the overall meaning of the context, so it is barely possible for TF-IDF to retrieve the appropriate questions. All the samples in Table \ref{tbrq1example} show that the retrieved questions are totally different from the original ones.

(3) BiLSTM-CC and BiLSTM outperform TF-IDF by a large margin, indicating the superiority of neural generative models. Besides, BiLSTM-CC outperforms the vanilla BiLSTM by 11\% on average on the joint dataset and by 29\% on average on separated subsets, which proves the effectiveness of the copy and coverage mechanisms. From the samples in Table \ref{tbrq1example}, we can find that BiLSTM often borrows the exact phrases from question bodies, while BiLSTM-CC can reorganize words into sentences.
Despite the good performance of BiLSTM-CC, our CCBERT model outperforms it by 9\% on average on the joint dataset and by 11\% on average on separated subsets, indicating the superiority of Transformer-based models and the pre-training strategy. From the generated samples in Table \ref{tbrq1example}, we can find that CCBERT is better at handling long-range dependencies than BiLSTM-CC. For instance, in the first sample, CCBERT notices that "this shape" refers to the "TriangleMesh" that appeared later in the question body, while BiLSTM-CC tends to focus on the content at the beginning of the question body. Furthermore, it is the same for the rest of the samples, where unwanted words (i.e., "android studio", "spring boot", and "crypto-stock") in the front of the question body attract more attention from BiLSTM-CC. At the same time, CCBERT can find the critical words (i.e., "sqlite", "service", and "key-value") that hide in the middle of the question body.

(4) BART is a competitive model, where CCBERT outperforms it by 1.3\% on average on the joint dataset and by 3.6\% on average on separated subsets. According to the samples in Table \ref{tbrq1example}, BART is good at generating clear and readable titles because it is a generation-oriented model that has been pre-trained on a vast natural language corpus. However, we can see from the first and second samples that titles generated by BART miss the keywords (i.e., "TriangleMesh" and "sqlite"), we attribute this problem to BART's inferior understanding of source code. For instance, in the first sample, BART cannot find that the "shape" at the beginning refers to the "TiangleMesh" object declared in the following code snippet. In the second sample, a major part of the body describes inserting data into the SQLite database, while BART only focuses on the unimportant word "android studio". On the contrary, with the help of bi-modal pre-trained CodeBERT encoder, our CCBERT model better understands the source code and generates more semantic titles relevant to the original ones.

(5) The Oracle model has a surprisingly good performance on both subsets, which shows much space for improving current models. In terms of the recall-oriented ROUGE metric, the excellent performance of the Oracle model indicates that most tokens in a question title come from the corresponding question body. However, our CCBERT model can only identify a part of the useful tokens in question bodies, leading to a moderate performance on the BLEU metric. Nevertheless, we can find that all the titles generated by the Oracle model hardly adapt to the grammatical norms from the generated samples in Table \ref{tbrq1example}, which indicates the necessity of applying generative models on this task. As for the reasons of the huge performance difference between our CCBERT model and the Oracle model, we think that our model may not have well handled the complex long-range bi-modal contexts, and the personalized writing habits of question titles also makes it hard for an automatic model to summarize in the same way as developers do.

\begin{tcolorbox}
	Answer to RQ-1: Our CCBERT model outperforms the TF-IDF, BiLSTM, BiLSTM-CC, and BART models regarding all the automated evaluation metrics on both training strategies.
\end{tcolorbox}

\subsection{RQ-2: What is the advantage of using the bi-modal information of the entire question body? }
\label{rq2}

\begin{table*}
	\centering
	\caption{The performance of CCBERT and BiLSTM-CC on the datasets with different sizes. All the score numbers are averages over the tested posts of different languages.}
	\label{tbrq3}
	\resizebox{0.65\textwidth}{!}{
	\begin{tabular}{@{\ }cccccc@{\ }}
		\toprule Model
		& Language                  & BLEUS-4     &  ROUGE-1    &   ROUGE-2  &  ROUGE-L    \\ \midrule[0.6pt]
		\multirow{4}{*}{BiLSTM-$\mathrm{CC_{joint}}$}
		& \scriptsize Java    & 19.73 & 41.10 & 19.54 & 40.04  \\
		& \scriptsize Python    & 19.74 & 42.67 & 19.97 & 41.72  \\
		& \scriptsize JS &  20.59 & 42.62 & 20.36 & 41.59  \\
		& \scriptsize PHP    & 20.56 & 43.01 & 20.92 & 41.73 \\ \midrule[0.1pt]
		\multirow{4}{*}{BiSLTM-$\mathrm{CC_{joint/2}}$}
        & \scriptsize Java & 18.99 & 39.92 & 18.05 & 39.20 \\
        & \scriptsize Python & 19.52 & 42.00 & 19.36 & 41.40 \\
        & \scriptsize JS & 19.58 & 41.21 & 18.76 & 40.50 \\
        & \scriptsize PHP & 20.06 & 41.83 & 20.03 & 41.05 \\ \midrule[0.1pt]
		\multirow{4}{*}{BiSLTM-$\mathrm{CC_{joint/4}}$}
        & \scriptsize Java & 18.74 & 39.73 & 17.81 & 39.09 \\
        & \scriptsize Python & 19.28 & 41.93 & 18.82 & 41.11 \\
        & \scriptsize JS & 19.43 & 40.83 & 18.41 & 40.06 \\
        & \scriptsize PHP & 20.02 & 41.39 & 19.70 & 40.83 \\
		\midrule[0.1pt]
		\multirow{4}{*}{BiSLTM-$\mathrm{CC_{joint/8}}$}
        & \scriptsize Java & 17.34 & 38.14 & 16.50 & 37.64 \\
        & \scriptsize Python & 17.61 & 39.98 & 17.12 & 39.52 \\
        & \scriptsize JS & 18.39 & 39.60 & 17.32 & 39.16 \\
        & \scriptsize PHP & 18.72 & 40.34 & 18.20 & 39.56 \\ \midrule[0.6pt]
		\multirow{4}{*}{$\mathrm{CCBERT_{joint}}$}
		& \scriptsize Java    & 21.16 & 44.26 & 21.58 & 42.92  \\
		& \scriptsize Python    & 22.40 & 46.88 & 22.89 & 44.92  \\
		& \scriptsize JS    & 22.18 & 45.72 & 22.40 & 44.15  \\
		& \scriptsize PHP    & 22.65 & 47.03 & 23.50 & 45.15  \\ \midrule[0.1pt]
		\multirow{4}{*}{$\mathrm{CCBERT_{joint/2}}$}
        & \scriptsize Java & 21.02 & 43.78 & 20.78 & 42.08 \\
        & \scriptsize Python & 21.83 & 45.84 & 22.21 & 44.14 \\
        & \scriptsize JS & 21.89 & 44.75 & 21.79 & 42.85 \\
        & \scriptsize PHP & 22.64 & 46.21 & 23.24 & 44.27 \\ \midrule[0.1pt]
		\multirow{4}{*}{$\mathrm{CCBERT_{joint/4}}$}
        & \scriptsize Java & 20.56 & 43.45 & 20.52 & 41.83 \\
        & \scriptsize Python & 21.22 & 45.81 & 21.72 & 44.02 \\
        & \scriptsize JS & 21.55 & 44.68 & 21.45 & 42.76 \\
        & \scriptsize PHP & 22.15 & 46.06 & 22.54 & 44.19 \\ \midrule[0.1pt]
		\multirow{4}{*}{$\mathrm{CCBERT_{joint/8}}$}
        & \scriptsize Java & 20.45 & 43.06 & 20.17 & 41.45 \\
        & \scriptsize Python & 20.73 & 44.42 & 20.92 & 43.56 \\
        & \scriptsize JS & 20.89 & 43.44 & 20.36 & 41.75 \\
        & \scriptsize PHP & 21.95 & 45.12 & 22.26 & 43.56 \\ \bottomrule
	\end{tabular}
	}
\end{table*}

\noindent \textbf{Motivation: }
Although we have illustrated the necessity of using both text descriptions and code snippets to generate high-quality question titles, we would like to quantify the improvement of using the bi-modal information over the code-only setting in Gao et al.'s work \cite{gao2020generating}.

\noindent \textbf{Method: }We post-process all question bodies in $Data_{exp}$ to keep code snippets and weed out text descriptions. We follow the jointly training strategy in this experiment. For the convenience of comparison, the new code-only dataset has questions in the same order as the previous joint dataset during training and testing. We choose BiLSTM-CC and CCBERT as representatives of our generative models and show their performance in Table \ref{tbl4}.

\noindent \textbf{Results: } There is a severe decline in the performance of both models when using only code snippets for training.
Specifically, the performance of CCBERT declines by 37\% on average, and BiLSTM-CC drops its performance by 36\% on average. Such results are expected because code snippets themselves can not offer sufficient context to a question. 

According to the samples in Table \ref{tbrq1example}, it is hard to tell the corresponding titles of all the four samples given only code snippets. Therefore, the generated titles may be incomplete and incorrect. For instance, in the first sample, both BiLSTM-CC$_{code}$ and CCBERT$_{code}$ pay attention to the "TriangleMesh", but neither of them deduces the word "shape" used in the title. In the second sample, both models fail to tell that the actual purpose of using "Android activity" and "SQLite database" is to insert user input data into the SQLite database. In the third and fourth samples, although both models manage to create the words (i.e., "using", "convert", "deserialize") that are not in the code snippets, there are still few overlaps between the generated words and the wanted ones.
However, under code-only circumstances and without changing model structure and hyperparameters, CCBERT$_{code}$ shows the superiority over BiLSTM-CC$_{code}$, which also indicates its generalization ability on different tasks.

\begin{tcolorbox}
	Answer to RQ-2: Applying bi-modal information greatly boosts both models' performance, where CCBERT still outperforms BiLSTM-CC.
\end{tcolorbox}

\subsection{RQ-3: How effective is our CCBERT model under low-resource circumstances?}
\label{rq3}

\begin{table*}
	\centering
	\caption{
		The performance of CCBERT and BiLSTM-CC trained on $Data_{exp+}$ without the interrogative constraint. All the score numbers are averages over the tested posts of different languages.
	}\label{tbl_exp_plus_performance}
	\resizebox{0.65\textwidth}{!}{
	\begin{tabular}{@{\ }cccccc@{\ }}
		\toprule
		Model                  & Language      & BLEUS-4 & ROUGE-1 & ROUGE-2 & ROUGE-L \\ \midrule[0.6pt]
		\multirow{4}{*}{BiSLTM-$\mathrm{CC_{exp+}}$}
        & \scriptsize Java & 16.71 & 31.79 & 14.52 & 30.36 \\
        & \scriptsize Python & 17.40 & 33.61 & 14.70 & 31.65 \\
        & \scriptsize JS & 17.78 & 33.67 & 15.38 & 32.16 \\
        & \scriptsize PHP & 18.27 & 34.32 & 15.46 & 32.19 \\
         \midrule[0.4pt]
		\multirow{4}{*}{$\mathrm{CCBERT_{exp+}}$}
        & \scriptsize Java & 18.68 & 36.32 & 16.43 & 33.81 \\
        & \scriptsize Python & 19.18 & 37.50 & 16.74 & 34.77 \\
        & \scriptsize JS & 19.55 & 37.81 & 17.04 & 35.31 \\
        & \scriptsize PHP & 20.19 & 39.02 & 17.05 & 35.69 \\ \bottomrule
	\end{tabular}
	}
\end{table*}

\noindent \textbf{Motivation: }Data-hungry is a common issue in the field of deep learning, which significantly hinders the application of many excellent models. Our model may need to train on massive high-quality questions. Since previous experiments have proved that CCBERT can handle the situation with around 200,000 samples for training, we carry out this experiment to discuss the effectiveness of our model under low-resource circumstances. 

\noindent \textbf{Method: }We first make several copies of $Data_{exp}$, and then randomly erase a certain amount of questions in the train sets, leaving the validation and test sets untouched. We choose three fractions as the percentage of samples to erase, which are 1/2, 3/4, and 7/8. This makes three new train sets sized of 98,909 ($\mathrm{Data_{exp}/2}$), 49,454 ($\mathrm{Data_{exp}/4}$), and 24,727 ($\mathrm{Data_{exp}/8}$). Along with the CCBERT model, we also train BiLSTM-CC on these datasets for comparison. Table \ref{tbrq3} presents the experimental results on the datasets with different sizes.

\noindent \textbf{Results: }It is as expected that both models have suffered performance degradation on smaller datasets. To our surprise, even if the amount of data decreases exponentially, the performance has a steady decline. 
Specifically, the performance of CCBERT declines by 2\% on average on the $\mathrm{Data_{exp}/2}$ subset, by 2.8\% on the $\mathrm{Data_{exp}/4}$ subset, and by 4.8\% on the $\mathrm{Data_{exp}/8}$ subset; while BiLSTM-CC drops its performance by 3\% on average on the $\mathrm{Data_{exp}/2}$ subset, by 3.8\% on the $\mathrm{Data_{exp}/4}$ subset, and by 8.2\% on the $\mathrm{Data_{exp}/8}$ subset.
It indicates that our task is not so sensitive to the data volume and further verifies the existence of a writing pattern shared by high-quality questions. Meanwhile, developers have personalized writing habits, so a more considerable amount of data can help eliminate such noise and improve the performance of our model. Facilitated by the pre-trained CodeBERT encoder, our CCBERT model is better initialized to resist data noise and requires fewer data. We can see from the results that with only one-eighth of the data, CCBERT still outperforms BiLSTM-CC trained on the full dataset.

\begin{tcolorbox}
	Answer to RQ-3: Compared to BiLSTM-CC, our CCBERT model shows significant superiority under low-resource circumstances.
\end{tcolorbox}

\subsection{RQ-4: How much influence does interrogative constraint have on model training?}
\label{rq4}

\noindent \textbf{Motivation: }Applying the interrogative constraint may reduce the data noise and make it easier to train our models. Nevertheless, our dataset could be narrowed and biased because of this. So we carry out this experiment to investigate the actual influence of interrogative constraint on our model's performance. \\
\noindent \textbf{Method: }We first build a dataset $Data_{exp+}$ in the same way as building $Data_{exp}$, except for applying the interrogative constraint. This way, we get a much larger dataset. Then, we train and evaluate both BiLSTM-CC and CCBERT models on the new dataset, following the jointly training strategy. The automated evaluation results are shown in Table \ref{tbl_exp_plus_performance}. The statistics of $Data_{exp+}$ is shown in Table \ref{tbl_exp_plus}.  \\
\begin{table}
	\caption{The partition size of $Data_{exp+}$}\label{tbl_exp_plus}
	\begin{tabular}{@{\ \ \ }cccc@{\ \ \ }}
		\toprule
		Language & Train & Validation & Test\\ \midrule
		Java & 183,443 & 7,000 & 7,000 \\
		Python & 207,323 & 7,000 & 7,000 \\
		JS & 174,374 & 7,000 & 7,000 \\
		PHP & 104,227 & 4,000 & 4,000\\ \midrule[0.2pt]
		Total & 669,367 & 25,000 & 25,000\\
		\bottomrule
	\end{tabular}
\end{table}
\noindent \textbf{Results: }We believe that a question's popularity does not necessarily attribute to a unified title format. Many other aspects like the clarity of question description and the popularity of the asked domain will influence the popularity of a question. 
We have manually studied the questions that we regard as high-quality ones and have no interrogatives in their titles. We find that those titles are always casually written. For example, a title can be a combination of keywords,\footnote{\url{https://stackoverflow.com/questions/53279561/java-month-enum}} a short phrase,\footnote{\url{https://stackoverflow.com/questions/53218222/capture-logs-in-a-test}} an error message,\footnote{\url{https://stackoverflow.com/questions/53344676/java-lang-illegalstateexception-inputstream-has-already-been-read-do-not-use}} etc. 

According to Table \ref{tbl_exp_plus_performance}, both models have poor performance on $Data_{exp+}$, despite having so much data for training. Specifically, the CCBERT model has an average 19.4\% lower evaluation score than using $Data_{exp}$, while it is 21.4\% for BiLSTM-CC. But in this experiment, our CCBERT model still outperforms BiLSTM-CC by 11.6\% on average, which verifies the superiority of our model under different circumstances. 

\begin{tcolorbox}
	Answer to RQ-4: Dropping the interrogative constraint will lead to a decline in the performance of both CCBERT and BiLSTM-CC models.
\end{tcolorbox}

\subsection{RQ-5: How effective is our CCBERT model under human evaluation?}
\label{rq5}

\begin{table*}
	\centering
	\caption{Human evaluation results of the TF-IDF, BiLSTM-CC, and CCBERT models trained on the joint dataset of four languages. The ratios of four different scores and the average score are listed grouped by different criteria.}
    \label{tbrq_human_result}
	\begin{tabular}{@{\ \ }ccccccc@{\ \ }}
		\toprule
		Criteria   & Model  & Score 1 & Score 2 & Score 3 & Score 4 & Avg Score\\ \midrule[0.6pt]
		\multirow{3}{*}{Readability}
		& \scriptsize TF-IDF    & - & 16.6\% & 81.2\% & 2.2\% & 2.856 \\
		& \scriptsize BiLSTM-$\mathrm{CC_{joint}}$    & - & 23.6\% & 76.2\% & 0.2\% & 2.766 \\
		& \scriptsize $\mathrm{CCBERT_{joint}}$ &  - & 19.2\% & 80.2\% & 0.6\% & 2.814 \\ \midrule[0.2pt]
		\multirow{3}{*}{Correlation}
		& \scriptsize TF-IDF    & 91.6\% & 8.4\% & - & - & 1.084 \\
		& \scriptsize BiLSTM-$\mathrm{CC_{joint}}$    & 27.6\% & 55.4\% & 14.2\% & 2.8\% & 1.922 \\
		& \scriptsize $\mathrm{CCBERT_{joint}}$    & 16.8\% & 47.4\% & 30.8\% & 5\% & 2.240 \\ \bottomrule
	\end{tabular}
\end{table*}

\noindent \textbf{Motivation: }Automated evaluation is not always trustworthy because it is hard to decide the actual human-perceived quality in different situations. The BLEU and ROUGE metrics used in this study mainly focus on the lexical ngram overlap between text sequences ignoring the grammatical correctness and the semantic similarity. So in this experiment, we will perform a more human-centered evaluation to investigate the overall quality of the titles generated by our models. \\
\noindent \textbf{Method: }We consider the two criteria introduced in Table \ref{tbrq_human} when manually evaluating the generated titles, either of them can be scored between 1 and 4. We randomly sample 500 questions in the test set of $Data_{exp}$ and then obtain 1,500 titles generated by the TF-IDF, BiLSTM-$\mathrm{CC_{joint}}$, and $\mathrm{CCBERT_{joint}}$ models. We invite five graduate students who are not co-authors to help us with this experiment. They are all experienced programmers and familiar with Stack Overflow. Each participant is assigned 100 questions, and we attach each question with three generated titles. The participants need to rate the titles based on the scoring standards manually, and they are blinded as to which title is generated by our model. The evaluation results are shown in Table \ref{tbrq_human_result}. \\
\noindent \textbf{Results: } In terms of the readability criteria, the performance of the three models is evenly matched. As expected, TF-IDF achieves the highest score because it retrieves and returns the titles written by developers. Our CCBERT model outperforms BiLSTM-CC by only 1.7\% and is only 1.5\% less good than TF-IDF.
To conclude, most of the generated titles by our CCBERT model are regarded as easy to read and understand. 

Regarding the correlation criteria, both the CCBERT and BiLSTM-CC models outperform TF-IDF by a large margin. It shows the superiority of generative models over the retrieval-based method on semantic understanding. The performance of our CCBERT model is 16.5\% better than BiLSTM-CC, which is not surprising because attributed to the pre-trained CodeBERT encoder, our model is more capable of handling long-range dependencies in bi-modal content. However, according to Table \ref{tbrq_human_result}, only less than a half of the generated titles of CCBERT are considered well matching the questions. It suggests that there is still much room for improvements in our approach. \\
\begin{tcolorbox}
	Answer to RQ-5: CCBERT performs much better than TF-IDF and BiLSTM-CC concerning the correlation criteria but a little bit worse than human written titles retrieved by TF-IDF concerning the readability criteria under human evaluation.
\end{tcolorbox}

\section{Related Work}
\label{related work}
Since we treat our TGEQB task as abstractive summarization, we take the related works in text summarization and code summarization for reference. We briefly introduce the recent literature in this section.

\subsection{Text Summarization}
There are extractive and abstractive ways for summarization tasks. Both ways have been attracting extensive research interest.

The extractive models select sentences or paragraphs from source texts to best match the target summary. The idea of using a hierarchical encoder and an extractor for document summarization was proposed by Cheng et al. \cite{Cheng2016NeuralSB}. 
Later, researchers have proposed various solutions to deal with different detailed problems. For example, Zhou et al. \cite{Zhou2018NeuralDS} argued that one should not separate the sentence scoring and selection steps, so they proposed an integrated model to merge the two steps. 
Xu et al. \cite{Xu2020DiscourseAwareNE} argued that BERT-based models could not capture dependencies among discourse units, which leads to the problem of having unwanted phrases in extracted summaries. To tackle this problem, they proposed to encode the rhetorical structure theory trees with a graph convolutional network.
Jia et al. \cite{Jia2020NeuralES} argued that BERT-based models neglect the inherent dependencies among reference sentences, and they proposed to refine the sentence representations with a redundancy aware graph attention network. 
These novel models performed well on semantic parsing. However, our task requires the model to give readable titles, where the extractive ways have been proved unworkable on our dataset (reference the Oracle method's performance in Table \ref{tbrq1example}).

In general, abstractive models are not restricted to selecting and rearranging the original text but to generating each word from a given vocabulary.
See et al. \cite{See2017GetTT} argued that vanilla attentional sequence-to-sequence models always produce inaccurate factual details and duplicate phrases. So they proposed to use a hybrid generator incorporating both the copy and coverage mechanisms. Gehrmann et al. \cite{Gehrmann2018BottomUpAS} found the problem that abstractive models were poor at content selection. Instead of adding fancy mechanisms, they proposed a two-stage process to train an extractor and then use it as bottom-up attention to guide the generator. Liu et al. \cite{Liu2019TextSW} extended this idea by using a two-stage fine-tuning on both extraction and generation tasks. In addition, they proposed to use different optimizers for the encoder and decoder to alleviate mismatch brought by different objectives. Lewis et al. \cite{Lewis2020BARTDS} proposed BART. This generative-oriented pre-training model has achieved excellent performance in abstractive summarization tasks, so we choose it as a baseline model to compare with ours.
Abstractive summarization is similar to our task in many aspects. However, our dataset is more challenging due to the complex bi-modal context and the difficult rare tokens, which is why we adopt the CodeBERT model and the copy mechanism.
\subsection{Code Summarization}
Code summarization aims to generate readable and meaningful comments that accurately describe the given programs or subroutines, which is very useful for code search and comprehension. 

One way to deal with source code is to treat it as a sequence. Iyer et al. \cite{Iyer2016SummarizingSC} first proposed to use attentional LSTMs to produce summaries describing code snippets, and they released their training corpus. Hu et al. \cite{Hu2018SummarizingSC} further looked into the possibility of using API knowledge to generate comments that better describe the functionality of source code. Wei et al. \cite{Wei2019RetrieveAR} proposed to use comments of existing similar source code to guide new comment generation. Wei et al. \cite{Wei2019CodeGA} exploited the relations between code summarization and code generation, and proposed a dual framework to train the two tasks simultaneously. The experimental results showed that performance improvements were achieved on both tasks. Hu et al. \cite{Hu2018DeepCC} argued that code tokens should not be processed sequentially. Hence, they proposed an abstract syntax tree-based structural code representation and verified its effectiveness in generating code comments. Ahmad et al. \cite{Ahmad2020ATA} first introduced the Transformer model to this task. They proposed to use a pairwise position encoding to capture the long-range dependencies among code tokens. 
The above approaches treated source code as text sequences, but they also valued the particular information hidden behind the code. Their experiments convinced us that the programming language is different from the natural language.

The other way is to convert the source code into other forms of representation. Wan et al. \cite{Wan2018ImprovingAS} used a sequential encoder as well as a tree-based encoder to capture the general information from code. They also applied an actor-critic network to overcome the exposure bias issue of the auto-regressive decoder. LeClair et al. \cite{LeClair2019ANM} also used dual encoders and incorporated the copy mechanism to reserve necessary tokens reported by the AST analyzer. Besides, they \cite{LeClair2020ImprovedCS} further proposed to use a graph-based neural architecture that achieved even better performance. Yang et al. \cite{DBLP:conf/iwpc/YangKYGWMZ21} employed a sequential encoder and a graph-based encoder to learn the global and local semantic information to generate code comments of smart contracts.
The studies mentioned above show that source code needs technological transformation for models to extract the semantic information for summarization. Unfortunately, we find that content marked with the "<code>" tag in our filtered SO questions are not always syntactically correct source code. Therefore, we treat the code snippets as a sequence of tokens and use CodeBERT as the encoder, which is code-aware and takes sequences as input.

\section{Threats to Validity}
\label{threats to validity}
In this section, we identify the potential threats that might affect the recurrence of our experiments and the validation of our results.

\textbf{The threats to internal validity} concern us in two aspects, one is the re-implementation of baselines, the other is the design of the CCBERT model. To address the first issue, we rebuild the default development environment and choose the recommended settings for baseline models. 
As for the second issue, we have made trade-offs between different techniques. For example, we give up using the coverage mechanism because it is incompatible with the parallel decoding fashion of our Transformer decoder. Further experiments also show that our model is not troubled by the repetition problem. Besides, training the large CCBERT model with insufficient data may cause the overfitting issue. To address this issue, our CodeBERT encoder has initialized its parameters through self-supervised learning with massive data in the pre-training stage. Furthermore, in our experiments, we train our model with a small learning rate, which also helps alleviate the problem of overfitting.

\textbf{The threats to external validity} primarily relate to the quality and generalizability of our dataset. We notice that the SOTorrent dataset proposed by Baltes et al. \cite{Baltes2018SOTorrentRA} shares a lot in common with ours. SOTorrent aims to provide access to the version history of SO content, which involves legacy formats issues and contains a lot of duplicate posts. We build our own dataset, because ours is directly extracted from the latest version of SO posts, which has a unified HTML-style format and can be easily parsed into text/code blocks by a naive HTML parser.
There may be a deviation between our dataset and the realistic data. To make our dataset more realistic, we build it concerning four programming languages, apply two training strategies for comparison, and choose the lately posted questions for testing.

\textbf{The threats to construct validity} mainly relate to the evaluation measures. Though the BLEU and ROUGE metrics have been widely used, automated evaluation is still an open problem in the domain of text generation \cite{Sellam2020BLEURTLR, Yeh2021ACA, Fabbri2021SummEvalRS}. We perform a human-centered evaluation in terms of the readability and correlation criteria to address this issue.

\section{Conclusion and Future Work}
\label{conclusion and future work}
In this paper, we propose a new task to summarize question titles from bi-modal context and a novel model named CCBERT to tackle this problem. CCBERT incorporates the copy mechanism and the CodeBERT model, which can handle rare tokens and capture the long-range dependencies between bi-modal tokens. We build a large-scale dataset with sufficient high-quality questions concerning four programming languages. We choose the BLEU and ROUGE metrics for automated evaluation and various baseline models for comparison. Both automated and human evaluation results demonstrate the superiority of our model. We have released our dataset and source code for follow-up researches.

For future work,  we will try to tackle the problem of handling very long sequences. In addition, we consider using Incremental Learning techniques to make our model continuously learn new knowledge from new samples and retain most of the knowledge already learned.

\section*{Acknowledgment}
This work is supported in part by the Natural Science Foundation of Chongqing City, China (No. cstc2021jcyj-msxmX1115), the Fundamental Research Funds for the Central Universities (WUT: 213110003 and 223110002), the project supported by Sanya Science and Education Innovation Park of Wuhan University of Technology (No. 2020KF0059), the General Research Fund of the Research Grants Council of Hong Kong (No. 11208017), the research funds of City University of Hong Kong (No. 7005028 and 7005217), and the Research Support Fund by Intel, USA (No. 9220097), and funding supports from other industry partners (No. 9678149, 9440227, 9229029, 9440180, and 9220103). We thank Guoping Nie for his helpful comments about this work.

\bibliographystyle{cas-model2-names}

\bibliography{references}

\end{document}